\documentclass{tlp}
\usepackage[english]{babel}

\usepackage{amsmath}
\let\proof\relax

\usepackage{amsthm}
\usepackage{amssymb}
\usepackage{graphicx}
\usepackage{url}
\usepackage{comment}
\usepackage{xspace}
\usepackage{subcaption}
\usepackage{todonotes}
\usepackage{booktabs}

\newcommand{\derives}{\ensuremath{\mbox{\,$:$--}\,}\xspace}

\newcommand{\naf}{\ensuremath{not\ }}

\usepackage{listings}
\usepackage{asplisting}
\lstnewenvironment{aspencoding}[1][]
{
	\lstset{
		language=asp,
		showstringspaces=false,
		formfeed=\newpage,
		tabsize=4,
		commentstyle=\color{asparagus},
		basicstyle=\ttfamily\scriptsize,
		keywordstyle=\color{blue},
		numbers=left,
		breaklines=true,
		literate={~} {$\sim$}{1},
		#1
	}
}
{
}

\newcommand{\lv}[1]{\lambda}

\newcommand{\asp}{\textsc{asp}\xspace}
\newcommand{\cav}{\textsc{cav}\xspace}
\newcommand{\sumo}{\textsc{sumo}\xspace}
\newcommand{\cavs}{\textsc{cav}s\xspace}
\newcommand{\vanet}{\textsc{vanet}\xspace}

\begin{document}

\lefttitle{Cardellini et al.}

\jnlPage{1}{8}
\jnlDoiYr{2021}
\doival{10.1017/xxxxx}

\title[Optimising Dynamic Traffic Distribution for Urban Networks with \asp]{Optimising Dynamic Traffic Distribution for Urban Networks with Answer Set Programming\thanks{Carmine Dodaro and Marco Maratea were supported by Italian Ministry of Research (MUR) under PNRR project FAIR ``Future AI Research", CUP H23C22000860006. Carmine Dodaro was supported by Italian Ministry of Research (MUR) under PNRR project Tech4You ``Technologies for climate change adaptation and quality of life improvement", CUP H23C22000370006; and by GNCS-INdAM. Mauro Vallati was supported by a UKRI Future Leaders Fellowship [grant number MR/T041196/1].}}

\begin{authgrp}
\author{\gn{Matteo} \sn{Cardellini}}
\affiliation{University of Genova, Italy and Politecnico of Turin, Italy\\matteo.cardellini@edu.unige.it}
\author{\gn{Carmine} \sn{Dodaro}}
\affiliation{University of Calabria, Italy\\carmine.dodaro@unical.it}
\author{\gn{Marco} \sn{Maratea}}
\affiliation{University of Calabria, Italy\\marco.maratea@unical.it}
\author{\gn{Mauro} \sn{Vallati}}
\affiliation{University of Huddersfield, UK\\
m.vallati@hud.ac.uk}
\end{authgrp}

\history{\sub{xx xx xxxx;} \rev{xx xx xxxx;} \acc{xx xx xxxx}}

\maketitle

\begin{abstract}
Answer Set Programming (\asp) has demonstrated its potential as an effective tool for concisely representing and reasoning about real-world problems. In this paper, we present an application in which \asp has been successfully used in the context of dynamic traffic distribution for urban networks, within  a more general framework devised for solving such a real-world problem. In particular, \asp has been employed for the computation of the ``optimal" routes for all the vehicles in the network. We also provide an empirical analysis of the performance of the whole framework, and of its part in which \asp is employed, on two European urban areas, which shows the viability of the framework and the contribution \asp can give.

\end{abstract}

\section{Introduction}

Avoiding congestion and controlling traffic in urban scenarios is becoming nowadays of utmost importance due to the rapid growth of our cities' population and vehicles. The effective control of urban traffic as a mean to mitigate congestion can be beneficial in an economic, environmental and health way. At the end of the 21st century, the world population is expected to increase to 10.9 Billion, adding well over 3 Billion people to the current population \citep{roser2013world}. This massive growth, which will directly translate in more vehicles roaming the streets of our cities, demands improvements in the transport infrastructure and a better utilisation of our roads for the purpose of avoiding congesting the network. Traffic jams have a negative impact on safety and fuel consumption, which directly translates to a higher cost for drivers and health issues for residents near highly trafficked roads, caused by bad air quality and noise pollution \citep{van2004driving}.  Artificial Intelligence techniques, based on automated planning, have already been employed for optimising traffic flow, and more general in transportation (see, e.g., \citep{cenamor2014planning,chrpa2016automated,vallati2016efficient,ramirez2018integrated,cardellini2021station,el2024pddl+}), with some benefits, but they fail to scale in the presence of large number of vehicles. 

In this paper, we present an application in which Answer Set Programming (\asp)  \citep{DBLP:journals/amai/Niemela99,baral2003,DBLP:journals/cacm/BrewkaET11,DBLP:journals/ngc/GelfondL91} has been successfully used in the context of dynamic traffic distribution for urban networks, within  a more general framework devised for solving such a real-world problem. The framework, which allows to efficiently optimise and simulate traffic flow in a large roads' network with hundreds of vehicles, is composed of four phases: network analysis, domain-independent search, route optimisation, and mobility simulation. Within the framework, \asp is employed for representing and reasoning about the optimisation of the flow of traffic inside a road network by finding the best combination (schedule) of routes for all the vehicles in the network.
  We have performed an analysis on real-world traffic data from two European urban areas in UK and Italy, utilising the state-of-the-art Urban Mobility Simulator \sumo \citep{SUMO2018} to keep track of the state of the network: the analysis tested the correctness of the solution, and proved the efficiency and capabilities of the presented solution to reduce the metrics considered, sometimes significantly. Moreover, it shows the contribution \asp gives in terms of performance and metrics: all instances of Milton Keynes and Bologna up to $600$ vehicles inside the network are solved, optimally, in a short time.

The paper is structured as follows. Section \ref{sec:back} presents preliminaries about \asp. Then, Section \ref{sec:case} introduces the problem and the solution framework. The last two phases of the framework, i.e., the optimisation via \asp and the mobility simulator, are presented in Section \ref{sec:opt} and Section \ref{sec:exp}, respectively, together with the experiments we performed on real data. The paper ends in Section \ref{sec:related} and \ref{sec:conc} by discussing related work and by drawing some conclusions, respectively. 

\section{Background on ASP}
\label{sec:back}

Answer Set Programming (\asp) is a programming paradigm
developed in the field of non-monotonic reasoning and logic programming.
In this section, we overview the language of \asp. %
More detailed descriptions and a more formal account of \asp, including
the features of the language employed in this paper, can be found
in~\citep{DBLP:journals/cacm/BrewkaET11,DBLP:journals/tplp/CalimeriFGIKKLM20}.
Hereafter, we assume the reader is familiar with logic programming conventions.

\paragraph{\bf Syntax.}
Variables are strings starting with an uppercase letter, and
constants are non-negative integers or strings starting with lowercase letters.
A {\em term} is either a variable or a constant.
A {\em standard atom} is an expression $p(t_1, \ldots, t_n)$, where $p$ is a
{\em predicate} of arity $n$ and $t_1, \ldots, t_n$ are terms.
An atom $p(t_1, \ldots, t_n)$ is ground if $t_1, \ldots, t_n$ are constants.
A {\em ground set} is a set of pairs of the form $\langle consts\! :\!conj \rangle$,
where $consts$ is a list of constants and $conj$ is a conjunction of ground standard atoms.
A {\em symbolic set} is a set specified syntactically as
$\{\mathit{Terms}_1 : \mathit{Conj}_1; \cdots; \mathit{Terms}_t : \mathit{Conj}_t \}$,
where $t>0$, and for all $i \in [1,t]$, each $\mathit{Terms}_i$ is a list of terms such that $|\mathit{Terms}_i| = k > 0$, and  each $\mathit{Conj}_i$ is a conjunction of standard atoms.
A {\em set term} is either a symbolic set or a ground set.
Intuitively, a set term $\{X\! :\! a(X,c), p(X);Y\! :\! b(Y,m)\}$
stands for the union of two sets: the first one contains the $X$-values making the conjunction $a(X,c), p(X)$ true, and the second one contains the $Y$-values making the conjunction $b(Y,m)$ true.
An {\em aggregate function} is of the form $f(S)$, where $S$ is a
set term, and $f \in \{\#count, \#sum\}$ is an {\em aggregate function symbol}.
An {\em aggregate atom} is of the form $f(S) \prec T$, where $f(S)$ is an
aggregate function, $\prec\ \in \{<, \leq, >, \geq, \neq, =\}$
is an  operator, and $T$ is a term called guard.
An aggregate atom $f(S) \prec T$ is ground if $T$ is a constant and
$S$ is a ground set.
An \emph{atom} is either a standard atom or an aggregate atom.
A {\em rule} $r$ has the following form:

\begin{center}
	$a_1 \ | \ \ldots \ | \ a_n \ \derives \ b_1,\ldots, b_k, \naf b_{k+1},\ldots, \naf b_m.$
\end{center}

\noindent where $a_1,\ldots ,a_n$ are standard atoms, $b_1,\ldots ,b_k$ are atoms,
$b_{k+1},\ldots ,b_m$ are standard atoms, and $n,k,m\geq 0$.
A literal is either a standard atom $a$ or its negation $\naf a$.
The disjunction $a_1 \ | \ \ldots  \ | \ a_n$ is the {\em head} of $r$, while
the conjunction $b_1 , \ldots,  b_k, \naf b_{k+1} ,$ $\ldots, \naf b_m$ is its {\em body}. Rules with empty body and with only one atom in the head (i.e., $n=1$) are called {\em facts}.
Rules with empty head are called {\em constraints}. %
A variable that appears uniquely in set terms of a rule $r$ is said to be {\em local} in $r$, otherwise it is a {\em global} variable of $r$.
An \asp program is a set of {\em safe} rules, where
a rule $r$ is {\em safe} if the following conditions hold:
{\em (i)} for each global variable $X$ of $r$ there is a positive standard atom
$\ell$ in the body of $r$ such that $X$ appears in $\ell$, and
{\em (ii)} each local variable of $r$ appearing in a symbolic set
$\{ \mathit{Terms}\! :\! \mathit{Conj}\}$ also appears in a positive atom in $\mathit{Conj}$.
A {\em weak constraint}~\citep{DBLP:journals/tkde/BuccafurriLR00} $\omega$ is of the form:

\begin{center}
	$:\sim b_1,\ldots, b_k, \naf b_{k+1},\ldots, \naf b_m.\ [w@l] $
\end{center}

\noindent where $w$ and $l$ are the weight and level of $\omega$, respectively.
(Intuitively, $[w@l]$ is read as ``weight $w$ at level $l$'', 
where the weight is the ``cost'' of violating the condition in the body of $w$, 
whereas levels can be specified for defining a priority among preference criteria).
An \asp program with weak constraints is $\Pi = \langle P,W \rangle$, where $P$ is a program
and $W$ is a set of weak constraints.
A standard atom, a literal, a rule, a program or a weak constraint is {\em ground} if no variables appear in it.

\paragraph{\bf Semantics.}
Let $P$ be an \asp program. The {\em Herbrand universe} $U_{P}$ and
the {\em Herbrand base} $B _{P}$ of $P$ are defined as usual.
The ground instantiation $G_P$ of $P$ is the set of all the ground instances of rules of $P$ that can be obtained by substituting variables with constants from $U_{P}$.
An {\em interpretation} $I$ for $P$ is a subset $I$ of $B_{P}$.
A ground literal $\ell$ (resp., $\naf \ell$) is true w.r.t. $I$
if $\ell \in I$ (resp., $\ell \not\in I$), and false (resp., true) otherwise.
An aggregate atom is true w.r.t. $I$ if the evaluation of its aggregate function
(i.e., the result of the application of $f$ on $S$) w.r.t. $I$
satisfies the guard; otherwise, it is false.
A ground rule $r$ is {\em satisfied} by $I$
if at least one atom in the head is true w.r.t. $I$ whenever all conjuncts of the body
of $r$ are true w.r.t. $I$.
A model is an interpretation that satisfies all rules of a program.
Given a ground program  $G_P$ and an interpretation  $I$, the
{\em reduct} \citep{DBLP:journals/ai/FaberPL11} of $G_P$ w.r.t. $I$ is the subset $G_P^I$ of $G_P$ obtained
by deleting from $G_P$ the rules in which a body literal is false w.r.t. $I$.
An interpretation $I$ for $P$ is an {\em answer set} (or stable model)
for  $P$ if $I$ is a minimal model (under subset inclusion) of $G_P^I$
(i.e.,  $I$ is a minimal model for $G_P^I$) \citep{DBLP:journals/ai/FaberPL11}.
Given a program with weak constraints $\Pi = \langle P,W \rangle$, the semantics of $\Pi$ extends from the basic case defined above. Thus, let $G_{\Pi} = \langle G_P,G_W \rangle$ be the instantiation  of $\Pi$; a constraint $\omega \in G_W$ is violated by an interpretation $I$ if all the literals in $\omega$ are true w.r.t. $I$.
An {\em optimum answer set} for $\Pi$ is an answer set of $G_P$  that minimises
the sum of the weights of the violated weak constraints in $G_W$ in a prioritised way.

\paragraph{\bf Syntactic shortcuts.}
In the following, we also use \textit{choice rules} of the form $\{p\}$, where $p$ is an atom. Choice rules can be viewed as a syntactic shortcut for the rule $p\ | \ p'$, where $p'$ is a fresh new atom not appearing elsewhere in the program, meaning that the atom $p$ can be chosen as true.

\section{Case Study and Proposed Solution Framework}
\label{sec:case}

In this section, we present our case study about dynamic traffic distribution  for urban networks, and the solution framework we propose, in two separate subsections.

\subsection{Problem Description}\label{sec:problem-description}

Urban traffic routing aims to mitigate traffic congestion by navigating the vehicles and recommending less-congested routes, hence supporting a better use of the capacity of the urban network. Recent advances in connected autonomous vehicles (\cavs) technology provide an opportunity for routing approaches to be increasingly practicable and to revolutionise the field, as the communication capabilities of \cavs can allow roadside agents to collect real-time traffic information, and can support real-time communication between the vehicle and a centralised traffic control system \citep{doi:10.1080/15472450.2017.1336053,DBLP:conf/itsc/VallatiC18}. A centralised approach aims to provide the optimal routes for each \cav from the perspective of the traffic management centre, with the clear benefit of having a holistic vision of the controlled urban region. In other words, the centralised approach allows considering both the dynamic system optimal  principle \citep{merchant1978model} and the dynamic user optimal principle \citep{friesz1989dynamic} when routing vehicles, and the trade-off between the two according to traffic conditions. 

In this work, we consider a centralised scenario where a controller oversees the network in real time. The controller has knowledge of the network structure, that can possibly be updated according to accidents or other unexpected events. While in operation, the controller is provided with the list of incoming vehicles ({\sl controlled}), and the position of the vehicles already navigating the network (hereinafter referred to as {\sl simulated}). On the basis of this information, the controller provides an optimised route for controlled vehicles, that aims at minimising overall congestion while considering the dynamic aspects of traffic. 

We assume that vehicles entering the controlled region communicate to the controller their destination and their current path, via an existing Vehicular Ad-hoc Network (\vanet) \citep{CUCOR20211312}, and the controller assesses the network status in terms of expected or recorded congestion, and returns a route to the vehicle under the form of a sequence of links to follow \citep{shahi2020mrgm}. The vehicle then follows the given path, and such path can be considered by the traffic controller for informing the routing of future incoming vehicles. 

It is worth highlighting that for the sake of traffic distribution and route optimisation, traffic signals are not explicitly modelled. This is established practice in the transportation field (see, e.g., \citep{BLIEMER2020303,BATISTA2021103076}) for a number of reasons. First, traffic signals can be implicitly modelled by considering the average time needed to navigate through a link and the corresponding junction to leave it: this is a very efficient way to reduce the complexity of the task. Second, the main driving factors of congestion in urban areas are demand, the structure of the network, and the capacity of links -- in a sense, traffic signals are only ensuring that shared resources of the network are accessed correctly. Further, the setting of traffic lights is often unknown, even for those working on fixed-time mode, while reactive traffic control approaches implement algorithms to react to perceived traffic conditions, so counter engineering their behaviour is not trivial and adds significant computational complexity to the traffic distribution task.

\subsection{The Solution Framework} \label{sec:framework-instantiation}

\begin{figure}[t!]
    \centering
    \includegraphics[width=1\textwidth]{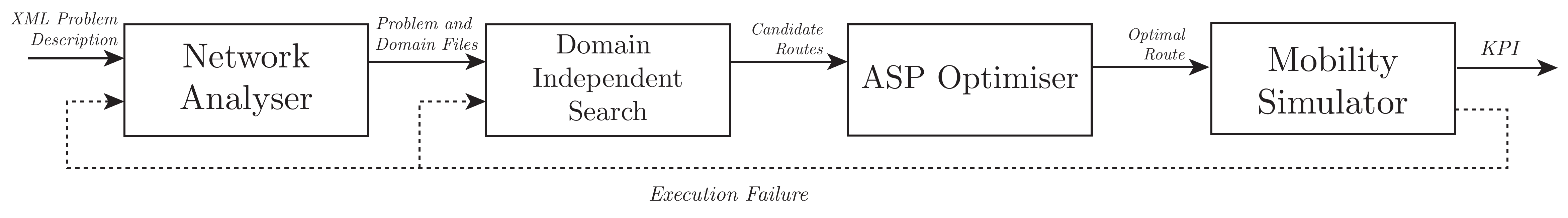}
    \caption{The solution framework.}
    \label{fig:instanced}
\end{figure}

Figure \ref{fig:instanced} presents our solution framework and its four components.
In the following, we describe the first two components, i.e., the Network Analyser and the Domain Independent Search, while the last two components will be presented in Sections~\ref{sec:opt} and~\ref{sec:exp}, respectively. The implemented framework, tailored to be used with the considered simulation and scenarios, is available at: \url{https://github.com/matteocarde/tplp-traffic}

\paragraph{\bf Network Analyser.} The purpose of the Network Analyser component is to simplify the topology of the network. It takes several inputs, including the network structure represented as a \textsc{xml} file, the list of incoming new vehicles, and the positions of vehicles that already have a route within the network. For the sake of this implementation of the framework, all input files are in the format used by the \sumo simulator. The Network Analyser then generates a simplified logical representation that describes the network, creating a model of the search space where viable solutions can be found in the next phase.

To achieve this, the Network Analyser employs a \textit{preprocessor} that is specifically designed to build an internal model of the road network. The preprocessor performs various tasks to simplify the network, for instance, $(i)$ joining small streets together, $(ii)$ simplifying intersections in roundabouts, and $(iii)$ removing streets from the network that cannot be used by vehicles, such as no-traffic zones or streets reserved for public transportation.  Regarding $(i)$ and $(ii)$, it is important to highlight a technical detail implemented by the processor. The output of the Network Analyser serves as input for the Domain Independent Search component and, subsequently, to the \asp Optimiser component. Since these components need to capture the complete flow of traffic, they must be time-dependent. To achieve accurate time-dependent analysis, it is crucial to discretise time into steps that are the right trade-off: being small enough to capture real traffic nuances but large enough to allow for efficient planning. The problem of finding the right discretisation step when planning in temporal scenarios is a well-known problem in planning (see, e.g.,~\citep{DBLP:journals/ai/GiganteMMS22,DBLP:conf/icaps/CardelliniMPSV24}). In our case, a large discretisation step could greatly impact on the quality of the produced plan. To exemplify this, let us suppose that a vehicle is about to navigate two streets, $s_1$ and $s_2$, with $s_1$ leading to $s_2$. The time to navigate $s_1$, at a speed of $45km/h$ is of $14s$, and the time for $s_2$ is of $12s$. In an ideal settings with constant speed and continuous time, it would take a vehicle $26s$ in total to cross the 2 streets. If instead we account for a discretisation step of $10s$, both $s_1$ and $s_2$ will take 2 discretisation steps to cross, i.e., $20s$, with a total time of $40s$. It is evident that large discretisation steps can lead to significantly misleading navigation times being considered. Further, with large discretisation steps, the times needed to cross streets tend to be the same, and thus the \asp Optimiser component could opt for solutions where the vehicles take routes with the less number of streets possible, but which in reality would amount to greater running times. Having a smaller discretisation step, like of $1s$, could guarantee that the solution is more cogent to the reality, but this would greatly increase the complexity of the optimisation problem and negatively affect the total solving time. 

Moreover, this discretisation step has an impact on the network's topology. For example, in most of the existing traffic simulation tools, a roundabout comprises multiple small streets that connect all incoming and outgoing streets at various intersections. However, representing each small street individually during the optimisation phase would not be accurate, as each of these small streets would need to be run in a time equal to the discretisation step, resulting in a large simulated running time not representative of the real running time.
To address this, the preprocessor combines the small streets that connect the incoming and outgoing streets of the roundabout, creating one longer street for each enter/exit combination of the roundabout, which can more realistically model the flow of traffic within the discretised time steps. As a result of this grouping, a single street is represented multiple times in the network. Ensuring that the total capacity of the roundabout is respected becomes the responsibility of the optimiser, which will ensure that the maximum number of vehicles inside the roundabout will be respected. 
Moreover, having removed any streets from the network that cannot be used by vehicles, such as no-traffic zones or streets reserved for public transportation, we might be left with intersections with only two intersecting streets and which can thus be joined to better deal with the discretisation step.
In our experimental settings, after having performed the aforementioned simplifications, we performed a simulation analysis on how different discretisation steps would have affected the running times of a vehicle that needed to run through several runs and choose the one that allowed for a total running time as close as possible to the real running time. We thus chose the discretisation step of $5s$.

\paragraph{\bf Domain Independent Search.} %
The Domain Independent Search component receives the simplified network representation generated by the Network Analyser as input. Its main purpose is to identify and output suitable routes for vehicles entering the network and determine the time ranges in which these vehicles will enter or exit each street along their routes.

The origin and destination of each approaching vehicle are known in advance, and the framework aims to find high-quality routes that connect these two points. To achieve this, the search phase computes all possible (acyclic) paths in the network graph that connect the source and target streets for each entering vehicle. However, in large and complex maps, this would result in an unmanageable number of routes. To tackle this, for each vehicle approaching the network, we firstly employ a Dijkstra search algorithm to explore the network graph. Starting from the source street, for each adjacent street, we define a new route, composed of the source street and the adjacent street, labelled with the sum of their length. We insert all the generated routes into a priority queue, we select the route with the smallest length and, by concatenating each adjacent street of the final street of the route with the route itself, we keep iterating the procedure, exploring the network. When the target street is reached, due to the properties of the Dijkstra algorithm, we know to have found the shortest route between the source and the target street. However, differently from the standard Dijkstra approach, we don't conclude the search when the target street is found, but we simply save the route, not putting it back in the priority queue and keep exploring the network, saving a new route every time the target street is reached. A loop detection mechanism is put in place to detect when the concatenation of a new street would cause a cycle and simply discard the generated route. We stop the procedure when a desired number of (acyclic) routes is reached (in our experimental setting, $60$ routes, appropriate for the sizes of the networks considered). These routes are guaranteed, by the Dijkstra algorithm, to be the shortest among all the possible routes but are not guaranteed to be the fastest, since congestion could change the running times of a street. 

Unfortunately, these routes, computed for each approaching vehicle, could still be too many to be considered by the solver. Moreover, most of the routes could only differ slightly by a few streets and thus quite interchangeable in distributing the traffic. Instead, we are interested in having a smaller number of routes but ``most different" between each other, i.e., sharing the least number of streets possible. This way, the \asp solver could decide to direct a vehicle to a longer route, but which is less congested and actually faster to run through. To compute these ``most different" routes for each pair of routes $r_1$ and $r_2$ we compute a similarity score through a function $\sigma(r_1, r_2) \in [0,1]$. In our experimental setting, the function $\sigma$ is computed as $$\sigma(r_1, r_2) = \frac{|\mathrm{streets}(r_1) \cap \mathrm{streets}(r_2)|}{\min(|\mathrm{streets}(r_1)|, |\mathrm{streets}(r_2)|)},$$ where $\mathrm{streets}(r)$ is the set of the streets of a route $r$. Then, we create a set $\mathcal{R} = \{R_1, \dots, R_q\}$ where each $R_i \in \mathcal{R}$ is a set of routes such that for each pair of routes $r_1, r_2$ we have $r_1,r_2 \in R_i$ if $\sigma(r_1, r_2) < \sigma_T$ and $r_1 \in R_i, r_2 \in R_j$ with $i\neq j$ if $\sigma(r_1, r_2) \geq \sigma_T$, with $\sigma_T \in [0,1]$ being a similarity threshold (in our experimental setting, $\sigma_T = 0.5$). The size $q$ of the set $\mathcal{R}$ depends on the instance but, usually, it is not greater than $3$, since for every source street being at the edge of the network, there are usually three cardinal directions in which the vehicles can go, and thus three possible sets of ``most different" routes. From each of the set $R_i$, we then select the top $k$ routes inside $R_i$, ordered by length (in our experimental setting, $k = 5$, thus usually having $15$ final routes to choose for each approaching vehicle). It is worth noting that the \asp Optimiser can therefore find the optimal route, according to the destination of the vehicle and the network condition, out of the identified $k$ final routes. In principle, the optimal route selected by the  \asp Optimiser can be different from the global optimal route: this is not deemed to be an issue in urban areas, where the highly dynamic nature of traffic plays a major role. %

We will see in the encoding of the \asp Optimiser in the following section that, to deal with the temporal dimension, we will discretise time by a quantum (in our setting $5s$) and then emulate the evolution of the movement of each vehicle running its route, where vehicles can enter or exit a street only on times multiple of this quantum. For the rapid execution of the \asp Optimiser component, we compute in advance time ranges for when vehicles could enter and exit the streets along their routes. This computation serves two purposes: speeding up the simulation of traffic flow and reducing overall computation time. By utilising the knowledge of all the vehicles' routes within the network (as determined when the vehicles are approaching the network), it becomes possible to compute time ranges for the entrance and exit timings in every street of a potential route run by an approaching vehicle by solving two relaxed simulations for each vehicle: 
\begin{enumerate}
    \item the first simulation is performed assuming that no other vehicle is present in the network, i.e., there is no congestion, and thus the vehicle can run at its maximum speed (45 km/h),
    \item the other simulation is performed assuming that all other vehicles in the map occupy all the streets of their routes at the same moment, thus obtaining, for each street, the maximum amount of congestion possible. For each street, we then compute the speed of the vehicles, proportional to the level of congestion, and from that speed we get the time to traverse the street. 
\end{enumerate}
 These two relaxed simulations provide an estimate of the minimum and maximum entry and exit time for each vehicle. Computing the speed proportional to the congestion can be done in several ways, in our experimental setting we perform it this way. We firstly compute the capacity of the street, which is the number of possible vehicles which could occupy the street. This number is proportional to the length of the street and the number of lines of the street. In our experimental settings, we make the assumption that a car occupies $8m$, i.e., the sum of the length of a standard car, usually $5m$, and the safety distance of $3m$ in the front. In a real scenario, the safety distance depends on the speed of the car, which is, however, what we want to compute. Nevertheless, in an urban scenario, where vehicles are not allowed high speed, the safety distance of $3m$ was validated by traffic operators. We compute the capacity of a street $s$ as \begin{equation}
     \mathrm{capacity}(s) = \left\lceil\frac{\mathrm{lines}(s) \times \mathrm{length}(s)}{8m} \right\rceil. \label{eq:capacity}
 \end{equation}
 We select the rounded up value to avoid small streets to have a capacity of $0$. We can now estimate the speed of the cars in the street $s$. Let $\mathrm{vehicles}(s)$ be the number of vehicles in the street $s$, i.e., its congestion. The speed of a street $s$ is estimated as
 \begin{equation}
     \mathrm{speed}(s) = \begin{cases}
         45km/h & \mathrm{if}\; 0 \leq \mathrm{vehicles}(s) < 0.4 \times \mathrm{capacity}(s),\\
         30km/h & \mathrm{if}\; 0.4 \times \mathrm{capacity}(s) \leq \mathrm{vehicles}(s) < 0.7 \times \mathrm{capacity}(s),\\
         15km/h & \mathrm{if}\; 0.7 \times \mathrm{capacity}(s) \leq \mathrm{vehicles}(s) .
     \end{cases}
     \label{eq:speed}
 \end{equation}
While this approach is only an estimation, it was validated, together with the thresholds (i.e., $0.4$ and $0.7$), by traffic operators and gave good results in the experimental setting.

The selected routes for each approaching vehicle together with the minimum and maximum entry and exit times are then passed to the third component of the architecture, the \asp Optimiser, dealing with optimisation of the routes.

\section{ASP for the Optimisation Phase}
\label{sec:opt}

In the section, we detail the third component of the solution framework, referred to as \asp Optimiser, which has the role to select the best route for each vehicle entering the network. 
Specifically, the input of the component is a set of \asp facts, referred to, in the following, as Data Model, and it uses an \asp encoding to compute the best route for every vehicle approaching the network.

\paragraph{\bf Data Model.} The data model consists of the following atoms:
\begin{itemize}
    \item \texttt{streetOnRoute(S,R,MIN,MAX)} represents the relationship between street \texttt{S} and route \texttt{R}. The variables \texttt{MIN} and \texttt{MAX}, computed by the Search component, define the expected time range for a vehicle starting at $t=0$ to traverse route \texttt{R} and enter street \texttt{S}.
    
    \item \texttt{link(S1,S2)} indicates that there is a connection from street \texttt{S1} to street \texttt{S2}.
    
    \item \texttt{vehicle(V,T)} defines the presence of a vehicle \texttt{V} of type \texttt{T} on the map. \texttt{T} can be either \texttt{con} (\emph{controlled} vehicle) or \texttt{sim} (\emph{simulated} vehicle). We remind that a controlled vehicle is a new vehicle entering the region for which a route has to be identified, while a simulated vehicle already has a set route, and the system only needs to track its presence on the map.
    
    \item \texttt{origin(V,FROM)} designates street \texttt{FROM} as the starting point of vehicle \texttt{V} during the planning process. For controlled vehicles, the origin represents the street where the vehicle first enters the map. Similarly, \texttt{destination(V,TO)} specifies street \texttt{TO} as the final destination that will take the vehicle \texttt{V} outside the map.
    
    \item \texttt{possibleRouteOfVehicle(V,R)} indicates that vehicle \texttt{V} can follow route \texttt{R} to move from its origin to its destination. For simulated vehicles, there will be only one possible route available, determined when the vehicle entered the network. For controlled vehicles, this atom represents a subset of all possible routes from the origin to the destination (as discussed in the previous section).
    
    \item \texttt{time(T)} represents the time unit used for scheduling, ranging from $0$ to the maximum horizon when all vehicles have left the network. %
    
    \item \texttt{capacity(S,N)} denotes the capacity of street \texttt{S}, with \texttt{N} indicating the capacity value, as computed by Eq. \ref{eq:capacity}.
    
    \item \texttt{trafficTravelTime(K,S,T)} represents the time required to traverse street \texttt{S} under different traffic conditions. The variable \texttt{K} can take values of \texttt{heavy}, \texttt{medium}, or \texttt{low}, indicating the traffic amount. These amounts are correlated to the speeds of 15 km/h, 30 km/h, or 45 km/h, respectively, as computed by Eq. \ref{eq:speed}. The time $\texttt{T}$ is simply computed by dividing the length of the street with the relative speed.
    
    \item \texttt{maxTrafficTravelTime(S,T)} models the time it would take to clear the street in case of congestion when it reaches its maximum capacity.
    
    \item \texttt{trafficThreshold(K,S,MIN,MAX)} defines the range between $\texttt{MIN}$ and $\texttt{MAX}$ of vehicles in street \texttt{S} that categorise $\texttt{K} \in \{\texttt{heavy}, \texttt{medium}, \texttt{low}\}$ traffic, corresponding to the traffic levels mentioned in Eq. \ref{eq:speed} (for example, if \texttt{K} is \texttt{medium} then $\texttt{MIN} = 0.4 \times \mathrm{capacity}(s)$ and $\texttt{MAX} = 0.7 \times \mathrm{capacity}(s)$).
    \item \texttt{roundabout(R,C)} indicates the existence of a roundabout \texttt{R} with a maximum capacity of \texttt{C}, computed as the sum of all capacities of the streets of the roundabout.
    
    \item \texttt{streetInRoundabout(SS,R)} indicates that the simplified (result of original street's grouping) street \texttt{SS} is part of the roundabout \texttt{R}.    
\end{itemize}

The output of the \asp Optimiser consists of the following atoms:
\begin{itemize}
    \item \texttt{solutionRoute(V,R)} assigns a route \texttt{R} to the vehicle \texttt{V}.
    \item \texttt{enter(V,S,IN)} and \texttt{exit(V,S,OUT)} specify the times when a simulated vehicle \texttt{V} enters and exits street \texttt{S}. We remind that an upper and lower bound of these values is determined in the Search phase and provided as input through the atom \texttt{streetOnRoute(.,.,.,.)}.
\end{itemize}
   
\begin{figure}[t!]
    \centering
    \begin{aspencoding}
$\label{enc:guessRoute1}$ {solutionRoute(V,R): possibleRouteOfVehicle(V,R)} = 1 :- vehicle(V,con).
$\label{enc:guessRoute0}$ solutionRoute(V,R) :- possibleRouteOfVehicle(V,R), vehicle(V,sim).
$\label{enc:guessEnter}$ {enter(V,S,T) : time(T), T=MIN..MAX} = 1 :- vehicle(V,con), solutionRoute(V,R), streetOnRoute(S,R,MIN,MAX), not origin(V,S).
$\label{enc:enterOrigin}$ enter(V,S,0) :- origin(V,S).
$\label{enc:guessExit}$ {exit(V,S,T) : time(T), T=IN+1..IN+MAX} = 1 :- vehicle(V,con), enter(V,S,IN), maxTrafficTravelTime(S,MAX).
$\label{enc:nVehicleOnStreet}$ nVehicleOnStreet(S,T,N) :- enter(_,S,T), N = #sum{1,V: enter(V,S,IN), IN <= T; -1,V: exit(V,S,OUT), OUT <= T}.
$\label{enc:travelTimeHeavy}$ travelTime(S,T,X) :- enter(_,S,T), nVehicleOnStreet(S,T,N), trafficThreshold(heavy,S,A,_), N >= A, trafficTravelTime(heavy,S,X).
$\label{enc:travelTimeMedium}$ travelTime(S,T,X) :- enter(_,S,T), nVehicleOnStreet(S,T,N), trafficThreshold(medium,S,A,B), N >= A, N < B, trafficTravelTime(medium,S,X).
$\label{enc:travelTimeLight}$ travelTime(S,T,X) :- enter(_,S,T), nVehicleOnStreet(S,T,N), trafficThreshold(low,S,_,B), N < B, trafficTravelTime(low,S,X).
$\label{enc:constrTravelTime}$ :- vehicle(V,con), exit(V,S,OUT), enter(V,S,IN), travelTime(S,IN,X), OUT < IN + X.
$\label{enc:constrLink}$ :- vehicle(V,con), exit(V,S1,OUT1), enter(V,S2,IN2), link(S1,S2), IN2 != OUT1.
$\label{enc:constrCapacity}$ :- enter(V,S,T), vehicle(V,con), capacity(S,MAX), nVehicleOnStreet(S,T,N), N > MAX.
$\label{enc:constrRoundabout}$ :- enter(V,SR,T), streetInRoundabout(SR,R), vehicle(V,_), roundabout(R,MAX), #sum{X,S: nVehicleOnStreet(S,T,X), streetInRoundabout(S,R)} > MAX.
$\label{enc:optCost}$ :~ nVehicleOnStreet(S,T,N). [N@2,S,T]
$\label{enc:optExit}$ :~ destination(V,S), exit(V,S,T). [T@1]
    \end{aspencoding}
    \caption{\asp Encoding used during optimisation.}
    \label{enc}
\end{figure}

\paragraph{\bf ASP Encoding.} 

Figure \ref{enc} presents the rules utilised in the \asp encoding. 
Rule $r_{\ref{enc:guessRoute1}}$ defines the atom \texttt{solutionRoute}, which, for every \textit{controlled} vehicle, selects exactly one route among the different selected routes (computed during the \textit{Search} phase) that transport the vehicle from its origin to its destination. On the other hand, rule $r_{\ref{enc:guessRoute0}}$ assigns the \texttt{solutionRoute} for \textit{simulated} vehicles to be the route they are already following.
Rule $r_{\ref{enc:guessEnter}}$ computes an entry time within the minimum and maximum range for each \textit{controlled} vehicle, as determined in the \textit{Search} phase. For the first street of origin, rule $r_{\ref{enc:enterOrigin}}$ enforces an entry time of zero. Similarly, rule $r_{\ref{enc:guessExit}}$ determines the exit time for vehicles at each street in their route.
The atom \texttt{nVehicleOnStreet(S,T,N)} is defined by rule $r_{\ref{enc:nVehicleOnStreet}}$ to count the number of vehicles on street \texttt{S} at time \texttt{T}. This atom is then used in rules $r_{\ref{enc:travelTimeHeavy}}$ to $r_{\ref{enc:travelTimeLight}}$ to compute the \texttt{travelTime(S,T,X)} atom, representing the time (\texttt{X}) required for a vehicle to traverse the entire length of street \texttt{S} under different traffic conditions: heavy, medium, or low. These computations assume that the vehicle enters street \texttt{S} at time \texttt{T}.
Constraint $r_{\ref{enc:constrTravelTime}}$ enforces that once a vehicle enters a street, it must remain on it for at least the amount of time specified by the \texttt{travelTime(.,.,.)} atom. This constraint serves as a lower-bound restriction, as the vehicle may remain congested and depart later.
Constraint $r_{\ref{enc:constrLink}}$ establishes an order among the streets within a route. Constraints $r_{\ref{enc:constrCapacity}}$ and $r_{\ref{enc:constrRoundabout}}$ ensure that vehicles adhere to street capacities and roundabout restrictions, respectively.
The weak constraint $r_{\ref{enc:optCost}}$ optimises the number of vehicles on the street. Then, as the second optimisation criteria, weak constraint $r_{\ref{enc:optExit}}$ imposes a preference for the solution that enables vehicles to reach their destinations more quickly.
The \asp encoding and some solution examples can be found in the following repository: \url{https://github.com/matteocarde/asp-traffic}.

\section{Experiments, Monitoring, and Execution} 
\label{sec:exp}

For the last phase of the framework (\textit{Mobility Simulator}), we have employed \sumo \citep{SUMO2018}, which is a state-of-the-art Urban Mobility Simulator to monitor the state of the network, execute the computed optimal solution plan, and verify its quality through the computation of Key Performance Indicators (KPIs), such as \textit{Total Duration} (i.e., the time the last vehicle exits the network), \textit{Average Route Length, Speed, Duration, Waiting Time} (i.e., the time vehicles spend in queues) and \textit{Depart Delay} (i.e., the time vehicles spend waiting for the road to free to enter the network). This phase takes the responsibility for executing the optimal plan found by the \asp Optimiser and to account for all the real-world (\textit{microscopic}) nuances not modelled in the previous phases. For example,  we leave to this component the job of simulating the flow of traffic light at intersections, cadenced by the switching phases of traffic lights. The overtakes among vehicles, the use of lanes, and the order in which vehicles are queued in traffic, are other aspects overlooked in the previous phases, in which the streets are considered as buckets with no particular order between them and the managed aspect is how much each vehicle will occupy (on average) the single street. 

Given the fact that the previous steps reason with an abstracted representation of the problem to solve, it is possible that the execution of the plan leads to a state of the world that is significantly different from the expected one. For instance, some roads expected to have free flows of vehicles can be very congested, and there can be vehicles queuing at junctions because of traffic not leaving the controlled region at the expected rate. These discrepancies can be provided as feedback to the Domain Independent Search, that can take them into account to define the current state of the network. Further, there may be disruptions that are modifying the viability of part of the network. For instance, a car accident can happen (they can be simulated in \sumo by modifying the behaviour of drivers), unexpected road works may be required, or extreme weather conditions can affect the area: this kind of events can reduce the capacity of roads, or completely block portions of the network, according to severity. This information about unexpected or unplanned events can be fed back to the Network Analyser to update its internal representation of the network and of the links, hence ensuring that the subsequent modules are reasoning upon a more accurate representation of reality. In this implementation we did not explicitly take into account these classes of events, as they are complex to simulate with \sumo and may require domain expertise to be properly modelled and encoded. However, the framework is designed to support this kind of updates, with no modification to the presented modules -- the implementation of the feedback loop in a target area is left for future work.

\begin{figure}[t!]
     \centering
     \begin{subfigure}{0.45\textwidth}
         \includegraphics[width=\textwidth]{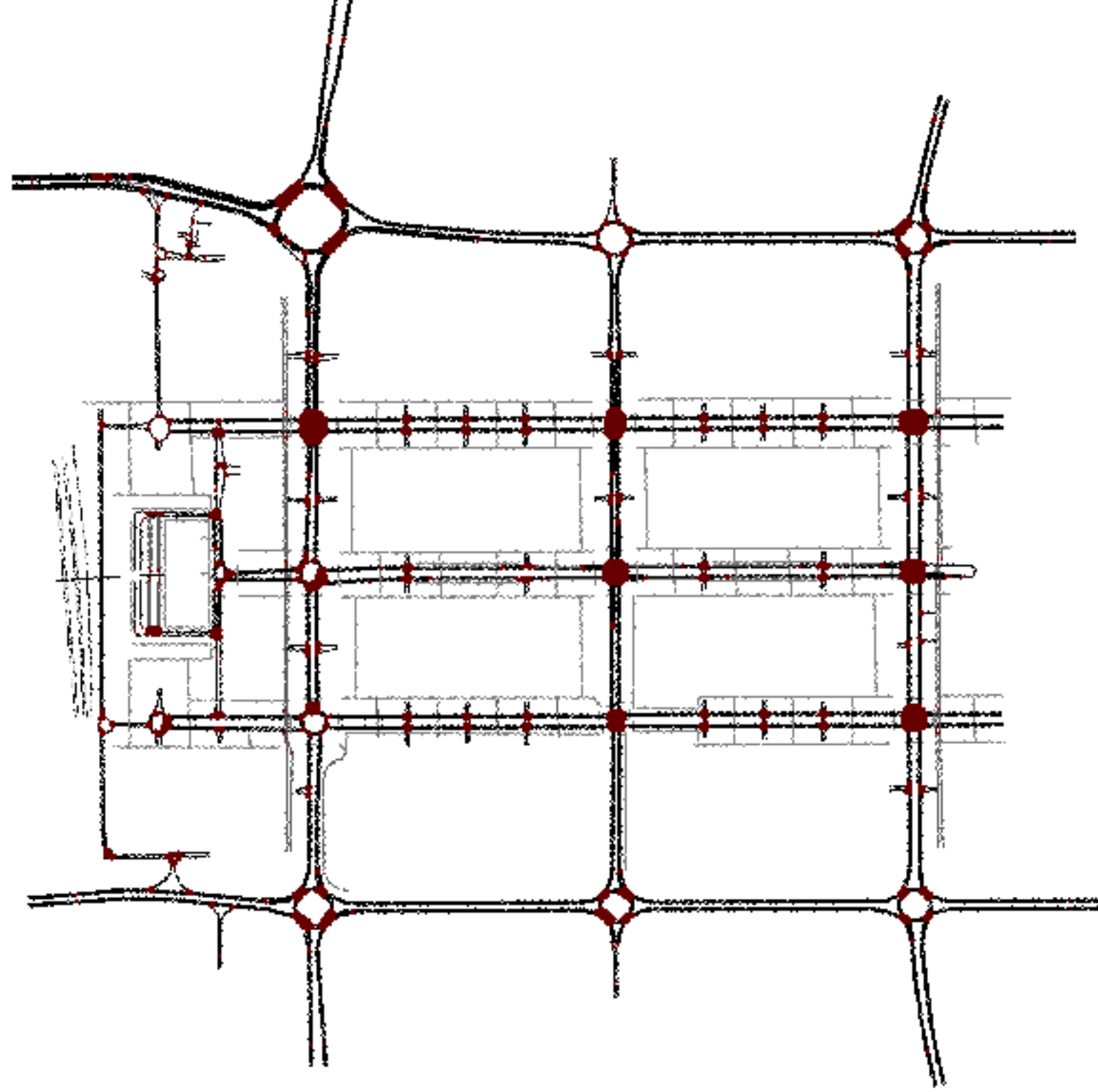}
         \caption{Milton Keynes}
         \label{fig:milton-keynes}
     \end{subfigure}
     \hfill
     \begin{subfigure}{0.45\textwidth}
         \centering
         \includegraphics[width=\textwidth]{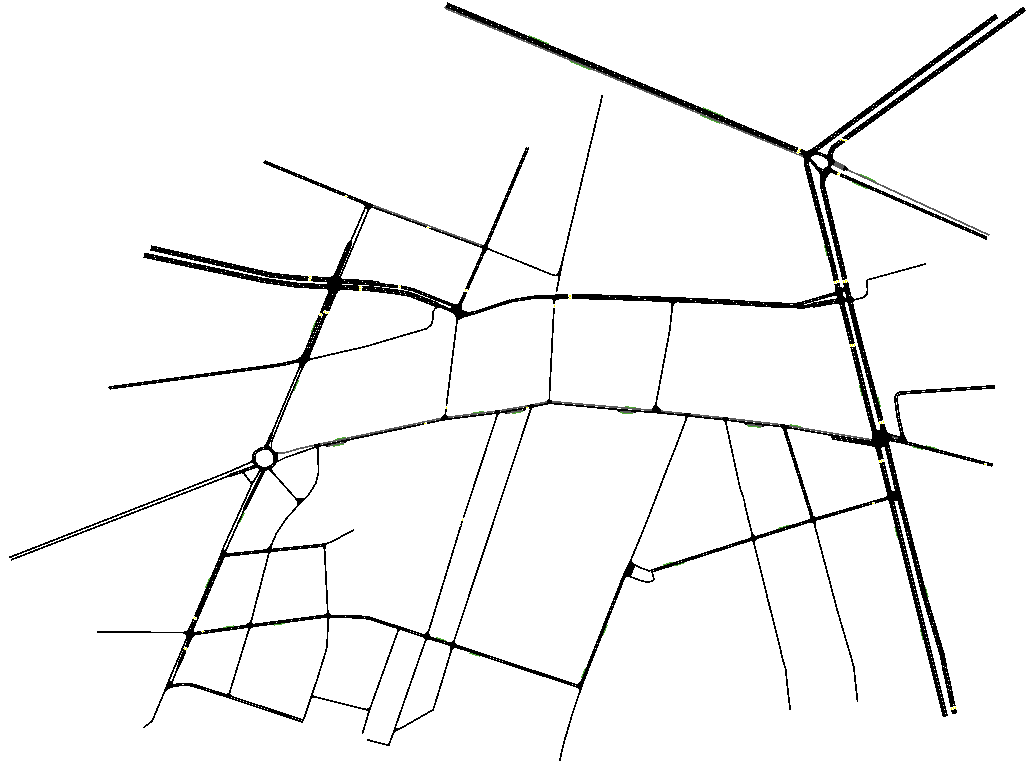}
         \caption{Bologna}
         \label{fig:bologna}
     \end{subfigure}
     \caption{The considered \sumo model of the central Milton Keynes (left) and Bologna (right) urban areas. Please note that the maps are not in scale, so can not be directly compared. } %
  \end{figure}

\subsection{Empirical Settings and Results}\label{sec:experiments}

We assess the proposed approach using \sumo and by considering two scenarios: Bologna and Milton Keynes. The Bologna \citep{dlr89354} scenario considers the $7^{th}$ city in Italy by population (approx. $400,000$). The considered area was constructed around the “Andrea Costa” road in Bologna, in which the football stadium is located. The network, represented in Fig. \ref{fig:bologna}, includes more than $110$ junctions and more than $170$ links. The total length of the modelled links is more than $33$ kilometres. The scenario includes the demand for Bologna’s peak hour (8am – 9am) in which $8,620$ vehicles roam the network, in some days of November 2008. The interested reader is referred to \citep{dlr89354} for a detailed description of how the network was constructed.

The Milton Keynes scenario considers the largest town in Buckinghamshire, United Kingdom, with a population of approximately $230,000$. A diagram of the considered network is shown in Fig. \ref{fig:milton-keynes}: it covers an area of approximately $2.9$ square kilometres. The network includes more than $25$ junctions and more than $50$ links. The total length of the modelled links is more than $45$ kilometres. The model simulates the morning rush hour, and has been built by considering historical traffic data collected between 8am and 9am on non-holiday weekdays. Data has been provided by the Milton Keynes Council, and gathered by sensors distributed in the region between December 2015 and December 2016. Traffic signal control information has been provided by the Council. The model has been calibrated and validated. During the morning rush hour, $1,900$ vehicles are entering the controlled region, and the main traffic flows are from North to South-East, and from West to East. This is because large residential areas are located at the North and West of the modelled region.

Our approach uses the TraCI interface\footnote{\url{https://sumo.dlr.de/docs/TraCI.html}} \citep{wegener2008traci} to interact with the \sumo simulation environment, to get the current network status, communicate with approaching vehicles, and inform vehicles of re-routing. Every time a new vehicle approaches the network, our approach is called, and a route is provided to the new entering vehicle. In both scenarios, the simulation is run until all the vehicles left the network. For each set of experiments, the simulation is run five times and results are averaged. All the experiments were run on a MacBook Pro with a 2.5 GHz Intel Core i7 quad-core, with 16 GB of RAM.

\begin{table}[t!]
\centering
\caption{Performance of actual traffic data coming from the Milton Keynes and Bologna's urban area, and the same vehicles routed using our proposed approach.}
\label{tab:empirical-results}
\begin{tabular}{@{\extracolsep{\fill}}lccccc}
  \topline
 & \multicolumn{2}{c}{Milton Keynes} & & \multicolumn{2}{c}{Bologna} \\
 \cmidrule{2-3}\cmidrule{5-6}
 & \multicolumn{1}{c}{Actual} & Our approach & & \multicolumn{1}{c}{Actual} & Our approach \\
 \cmidrule{2-2}\cmidrule{3-3} \cmidrule{5-5} \cmidrule{6-6}
Total Duration [s] & 15729 & 5065 && 5692 & 5647 \\
Avg. Route Length [m] & 2465 & 2107 && 1636 & 1678\\
Avg. Speed [m/s] & 2.49 & 5.28 && 6.37 & 6.58 \\
Avg. Duration [s] & 3718 & 515 && 283 & 281 \\
Avg. Waiting Time [s] & 3132 & 259 && 97 & 95 \\
Avg. Depart Delay [s] & 791 & 55 && 192 & 188
\botline
\end{tabular}
\end{table}

To empirically assess the performance of the proposed framework, in this subsection we compare actual simulated traffic data of the Milton Keynes and Bologna urban areas with a simulation in which the same vehicles are routed using our proposed approach, where $k$ has been set to $5$ as previously discussed. Table \ref{tab:empirical-results} shows a comparison between the two approaches in terms of a number of traditionally considered KPIs, that focus on delay, waiting time, speed, and path duration. As it can be seen from the comparison, in the Milton Keynes urban area, the proposed approach can greatly increase the overall performance of the network, spreading traffic and reducing congestion, increasing the average speed of vehicles and allowing the network to free faster. In Bologna, instead, the KPIs are only slightly increased with respect to the ones computed on actual traffic data, but still showing how our proposed approach can capture all the nuances of urban traffic control and to deal with the risk of congesting the network. The high differences in improvement which can be seen in Milton Keys with respect to Bologna can be explained by the two very different topologies of the networks. As it can be seen, in Milton Keynes (Figure \ref{fig:milton-keynes}) the streets form a sort of \textit{Manhattan Grid} in which parallel streets are more or less equal in terms of number of lanes, length, and intersections. For this reason, a car entering the network has a plethora of possible routes to choose, which are more or less of the same length, giving the possibility to better spread the traffic through the whole map. In Bologna (Figure \ref{fig:bologna}), instead, it can be noted how streets in the outer ring are more structured to deal with high volumes of traffic (with a large number of lanes and roundabouts to reduce traffic), while streets near the centre of the map (which constitute the residential area) have mostly one lane and many intersections. For this reason, vehicles have a smaller number of promising routes to choose from, leaving no choice to the planner but to let vehicles move through the streets of the outer ring. 

\subsection{ASP Evaluation}
\begin{figure}[t!]
     \centering
     \includegraphics[width=\textwidth]{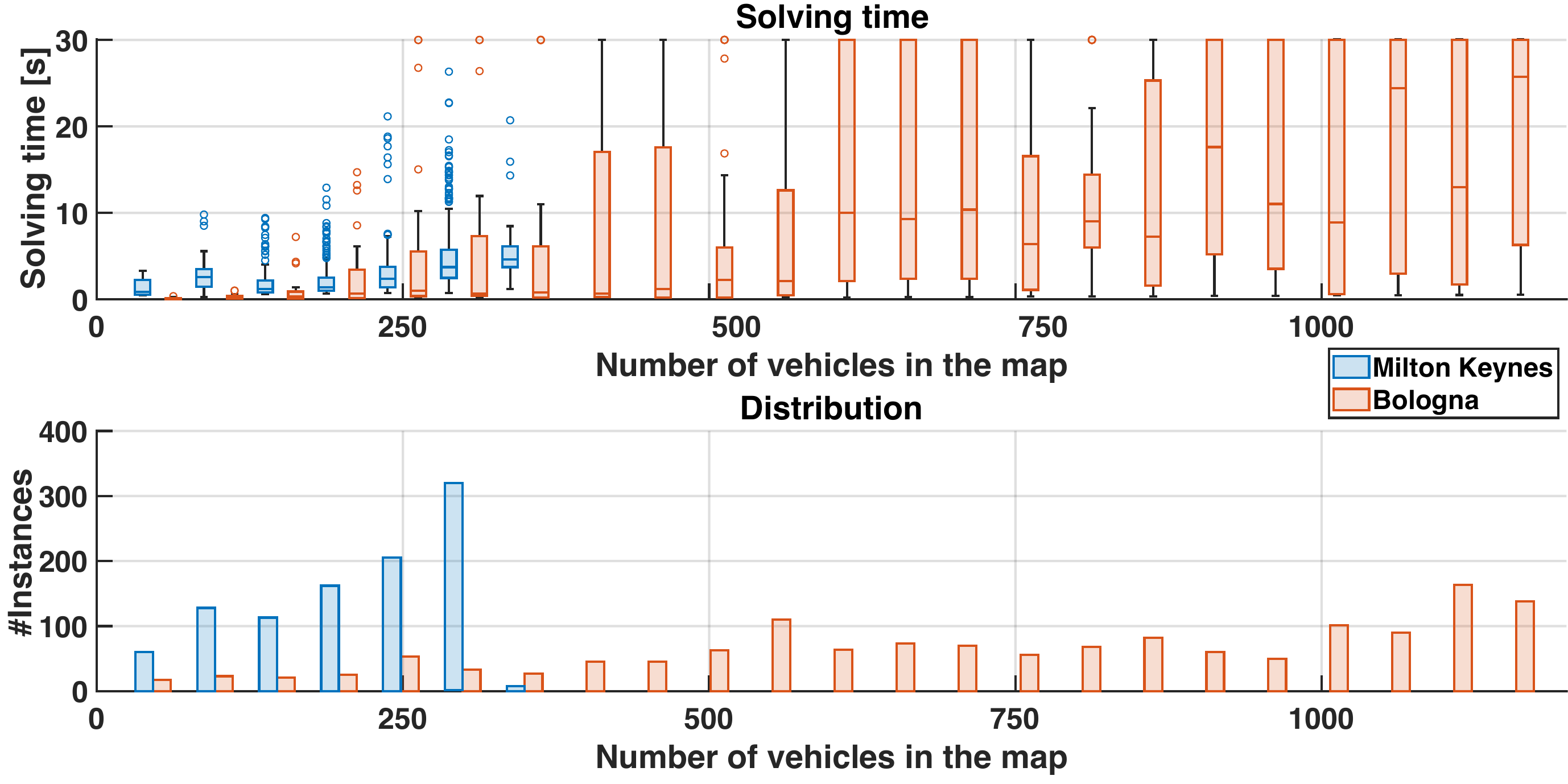}
     \caption{(Top) Boxplot of the solving time of \textsc{Clingo} in correlation with the number of vehicles inside the networks. (Bottom) Histogram representing the number of instances (i.e., each time a new vehicle enters the network) w.r.t. the number of vehicles inside the map. In the two charts, the x-axis has been clustered in bins of $50$.}
     \label{fig:scalability}
  \end{figure}

We now evaluate the \asp encoding presented in Figure \ref{enc} in terms of  capacity to rapidly find a high-quality solution to the routing of new vehicles inside the network.  Every time one vehicle approaches the network, the Network Analyser calls the Domain Independent Search, and then the \asp Optimiser is invoked to find optimal routes for the approaching vehicles. The discretisation step has been set to $5$ seconds according to some preliminary experiments. As \asp Optimiser, we employed \textsc{Clingo} configured with the option \texttt{--parallel-mode=2} which executes the task in two threads, one using the Branch and Bound algorithm \citep{DBLP:conf/lpnmr/GebserKK0S15}, and one using the Unsatisfiable Core algorithm \citep{DBLP:conf/iclp/AndresKMS12}. This approach has already been proven effective in other application domains, e.g.,  \citep{tplp_maugeri_2023}. Figure \ref{fig:scalability} presents the scalability results and dimensionality for the network of Milton Keynes and Bologna. %
It shows the correlation between the solving time of \textsc{Clingo} (Top) and the number of vehicles which roam the network (Bottom). The x-axis is grouped in bins of $50$, and the histogram below shows the number of instances (i.e., each time a new vehicle enters the network) for each bin. The instances of Milton Keynes and Bologna with less than $600$ vehicles can all be solved optimally within 30 seconds (actually, in the majority of cases, in less than $10$ seconds). When the dimensionality increases, the solver can still find optimal solutions in most of the cases (the median of the boxplot is always under 30 seconds), but for some instances the cut-off time is reached, and a suboptimal solution is returned. As shown in Table \ref{tab:empirical-results}, this approach can still improve the performances inside the networks.

\section{Related Work}
\label{sec:related}

One of the most common methods in the literature to optimise traffic flow in road networks is to schedule traffic signal switching phases, or \textit{signal phase plans}, with the aim of minimising vehicles' delay \citep{dotoli2006signal,papageorgiou2003review}. The well-known real time adaptive traffic control system \textsc{scoot} \citep{bretherton1990scoot}, for example, makes use of this methodology and is used extensively throughout Europe and the United Kingdom. Unfortunately, managing only the switching phases of traffic lights has some limitations: the only metric which is controllable and optimisable is the waiting time at intersections (which has a direct impact on the total travel time of vehicles) but is not straightforwardly expandable to consider other metrics (e.g., fuel consumption). Moreover, the (\textit{macroscopic}) point of view of traffic lights, which manages the flow of traffic modelling incoming and outgoing lanes as queues of vehicles, does not allow for a more detailed (\textit{microscopic}) consideration of the single vehicles and their routes inside the network. Microscopic simulation models have been largely discarded in the literature due to their high complexity and low scalability. In this paper, we leverage \asp, coupled with domain-dependent optimisations, to model the traffic flow of hundreds of vehicles inside large European cities from a \textit{microscopic} perspective.

The use of AI techniques in road transportation was shown to be effective in optimising traffic flows \citep{abduljabbar2019applications,miles2006potential}. For instance, \cite{vallati2016efficient} introduced a \textit{mixed discrete-continuous planning} \citep{fox2006modelling} approach for reducing congestion through traffic signal optimisation, that was subsequently improved to be deployed in urban areas of the United Kingdom \citep{mccluskey2017embedding,el2024pddl+}. %
\cite{chrpa2016automated}, instead, used a \textit{temporal planning} approach for managing traffic, through a \textit{microscopic} perspective, intending to reduce air pollution and respect air quality limitations. Other instances of urban traffic problems solved with automated planning technologies can be found in \cite{cenamor2014planning}. Even if automated planning has been beneficial in efficiently solving several real-world problems in transportation \citep{cardellini2021station,ramirez2018integrated}, the main point of failure is the ability to scale in the presence of large number of vehicles and the limited capabilities of generating optimised solutions, which are instead not issues for our approach. 

Noteworthy, \asp has already been used in this context. \cite{DBLP:conf/ecai/EiterFSS20} improve \textsc{scoot} for dynamically adjusting traffic signals via \asp. They employ \sumo for simulation, but do not rely on historical data. Our works focus on different aspects of traffic control, i.e., the distribution of connected vehicles. \cite{cardellini2023framework} proposed a framework to reason upon risks and their mitigation in the distribution of vehicles in urban areas. %
In railway, \cite{DBLP:journals/tplp/AbelsJOSTW21} deal with the train scheduling problem, %
with a hybrid solution as done by \cite{ramirez2018integrated}, which integrates \asp and difference logic, tested on real-world instances crafted by
domain experts at Swiss Federal Railways.
\cite{DBLP:conf/jelia/BeckEK12} use \asp to address the management of inconsistent traffic regulations, such as incorrect sign postings or software errors during data acquisition.

\section{Conclusion}
\label{sec:conc}

In this paper, we presented a framework for dealing with the problem of dynamic traffic distribution for urban networks. The framework is composed of four phases: Network Analyser, Domain Independent Search, \asp Optimiser and Mobility Simulator. The third phase of the approach relies on \asp for the computation of the optimal routes. The whole framework has been assessed on real-world data from two urban areas of the UK (Milton Keynes) and Italy (Bologna). Results show that the framework can significantly improve the considered traffic KPIs, and that the advantages  depend also on the topology of the networks. Further, the contribution \asp can give is evaluated in terms of performance and metrics: results outline that all instances of Milton Keynes and Bologna up to $600$ vehicles inside the network are solved, optimally, in a short time. 

We see several avenues for future work. First, we are interested in assessing the framework in different urban areas, and in enhancing the feedback capabilities by considering a number of critical failures and their impact. Second, we plan to extend the proposed approach to deal with vehicles that do not follow the provided instructions. 
Finally, we are interested in integrating the proposed framework into the larger urban traffic control infrastructure, so that information about calculated traffic flows can be exploited to inform traffic authorities policies and actions. %

\bibliographystyle{tlplike}
\bibliography{biblio}
\end{document}